# MemGEN: Memory is All You Need

Sylvain Gelly    Karol Kurach    Marcin Michalski    Xiaohua Zhai

Google Limbic System[*]
April 1st 2018

## Abstract

We propose a new learning paradigm called **Deep Memory**. It has the potential to completely revolutionize the Machine Learning field. Suprisingly, this paradigm has not been reinvented yet, unlike Deep Learning. At the core of this approach is the *Learning By Heart* principle, well studied in primary schools all over the world. Inspired by poem recitation, or by $\pi$ decimal memorization, we propose a concrete algorithm that mimics human behavior. We implement this paradigm on the task of generative modeling, and apply to images, natural language and even the $\pi$ decimals as long as one can print them as text. The proposed algorithm even generated this paper, in a one-shot learning setting. In carefully designed experiments, we show that the generated samples are indistinguishable from the training examples, as measured by any statistical tests or metrics.

## 1 Introduction

We follow the fundamental line of scientific research started by [LaLoudouana and Tarare, 2003], and later extended by [Albanie et al., 2017], and [Garfinkel et al., 2017]. Inspired by these approaches, we focus on the ultimate goal of generative modeling: outputing the same distribution as the input distribution. *Garbage In, Garbage Out* has been a well studied adage in Machine Learning, and our main contribution is to *literally* implement it.

---

[*]Part of Brain

Before we detail the theory (which nobody reads, but looks impressive) and the experiments (curves up-and-to-the-right, unless lower is better, in which case we transpose the figure), let's take a step back and think why do we want to do generative modeling? We can afford this pondering after a conference deadline because there is a lot of time before the next one, at least it seems so at the moment.

Many of us know that generative modeling is as good as a cake, and who does not like cakes? But let's pretend the cake is a lie for a moment. A Generative Model takes samples from a given data distribution and learns a model from it. We then hope that we can use this model to solve another task. By carefully studying [Shannon, 1948], one can notice that the best way to keep as much **information** as possible from the samples is to actually **store them** – full stop. After all, if nothing is lost, what do we lose?

Unfortunately, this idea alone is not sufficient. A key novel idea is to make use of advanced data structures like lists or even hashmaps[1] for generative modeling! Using those latest state-of-the-art technologies [Knuth, 1997], [Cormen, 2009], we are able to outperform old techniques from the 80s called Neural Networks which are no more than additions and multiplications, we have been told.

**Our main contributions:**
1. We propose a new learning paradigm. As an concrete application we show how to derive the ultimate generative modeling algorithm which *provably* outputs the *same distribution* as the input distribution.
2. Thanks to the algorithm efficiency on CPU, GPUs will be free to be used again for gaming [2]. The main downside of the computational efficiency is that PhD students can now compete with large organizations, which cannot leverage their more advanced infrastructure; this is unfair, as the infrastructure was hard to build.
3. Generated samples are statistically undistinguishable from real samples, we hence propose the ***Rademacher Coin Flipping*** metric which gives similar results more reliably. This finally closes the long debate on generative modeling metrics.

---

[1] These are often used in solutions to difficult interview questions, so they must be good.

[2] A similar approach can be taken to free GPUs from cryptocurrencies.

## 2  Related Work

[TODO(authors): add relevant papers here. For each, add some drawbacks, without necessarily reading them.]

Very relevant work probably happened before, but unfortunately we discovered it after writing the paper. What is important is to make sure that we cite well cited papers and famous authors. Hence, we prioritize by popularity and not by relevance. We of course did not forget to cite our previous papers, and those of our friends. That helps Search Engine Optimization [Beel et al., 2009].

[Placeholder for references requested by reviewers].

## 3  MemGEN: Remember It

We propose a specific implementation of MemGEN as illustration. This was initially written as pseudo-code but we then put our best software engineers on it, because code matters.

### 3.1  Algorithm

The details are presented in Algorithm 1. We stress that the algorithm is flexible and that one can apply more complicated data structures such as hashmaps.

---
**Algorithm 1:** *MemGEN*.

**Inputs :** Training data, of any type: $data$
**Function** `TrainModel`($data$):
  Initialize memory: $\ell \leftarrow []$
  *[Optionally do something useless which looks fancy to make the algorithm look more impressive]*
  **foreach** $x \in data$ **do**
    $\ell.append(x)$
  **return** $\ell$
**Function** `Sample`($\ell$):
  $i = randint(len(\ell))$
  **return** $\ell[i]$
TODO: Add test

---

## 3.2 Theory

It is always good to have some theory, even if it is not so relevant, as it looks nice and (as observed in [Graham et al., 1994]) you can impress your parents (and colleagues!) by just leaving *this* paper open on this page.

**Theorem 1** (Main result). *There exists an abelian and additive hyper-universal, non-multiply compact functional?*

*Proof.* The essential idea is that $\infty \in \mathcal{V}$. Let $\|\mathscr{P}\| \geq 1$. Because $\mathfrak{c} \neq \overline{-\infty^9}$, $f_{\mathbf{v},\delta} > 1$. So if $\pi$ is right-invariant and standard then

$$\tilde{\mathfrak{p}}\left(\sqrt{2}^9, q(H'')^4\right) \geq \bigcup_{K \in \pi} \overline{e_K^{-6}}.$$

Since there exists a super-hyperbolic, semi-Lie–Sylvester vector, $\mathbf{i}'' \geq B^{(B)}$.

Let $\hat{w}$ be an $\mathcal{A}$-Beltrami monodromy. Trivially, every $\hat{w}$-set is Cantor, hyper-compact, contra-Artinian and freely ultra-null. By smoothness, if $\mathbf{z}' \leq \|G\|$ then every differentiable $p$-adic function on $\mathcal{A}$ is analytic (in the sense of Tate's rigid analytic spaces). Since

$$\mathcal{S}(\mathcal{Z}^{(f)})i < \Phi^{(S)^{-1}}\left(\aleph_0 - 1\right) \wedge X'\left(\mathcal{F}^1, \ldots, \mathcal{F}^6\right) \wedge \cdots \wedge \cos\left(1\right)$$
$$> \frac{\mathscr{C}\left(-\sqrt{2}, -h'\right)}{-|\Gamma_u|},$$

if Artin's criterion applies then $\mathscr{S}' \cong \bar{I}(G)$. Trivially, $p \leq \bar{\mathbf{u}}(Q)$. This is a contradiction. Probably. □

The undisputed usefulness of the Theorem in the context of Deep Memory models is self-evident.

**Theorem 2** (Convergence). *MemGEN converges, in a good way, in $\Theta(N)$ steps with probability 1, where $N$ be the number of training examples.*

*Proof.* Follows directly from Theorem 1 and [de Lagrange, 1770][3].
Also, trivial when using the Borel $\sigma-$algebra on the sample space $\Omega$ and the measure $\mu$, especially when it is a Reproducing Kernel Hilbert Space (RKHS)[4].
□

---

[3] Actually this is unrelated, but it is nice to have an old reference, to ground the paper into history. And who reads references anyway?

[4] Even though the Deep Learning community seems to like them less.

# 4 Empirical Evaluation

To showcase our results, we will demonstrate how this simple method performs against state-of-the-art techniques. The current best metrics to evaluate generative models are based on distances between two samples, as well as human evaluations. We perform both to quantify the performance of our algorithm.

## 4.1 Results

### 4.1.1 Quantitative Results

**Distance from Two Samples:** Given a sample from the true distribution and a sample from the generated distribution, we measure how similar the two underlying distributions are. Obviously, we use the **Test Set** to represent the true distribution, as taking the **Training Set** would not reveal possible **overfitting** issues. The results are much better than any state-of-the-art result on all of the distance metrics. Figure 1 is illustrative of those results.

**Human Evaluation:** The raters are shown two images, one sampled from the model, and the other sampled from a holdout set, and they have to decide which one of the two looks better. Figure 2 shows the outcome of the human evaluation and demonstrates that humans cannot distinguish between true and fake samples. Actually, we could not differentiate those results with the results of unbiased coin flips. Hence, we now propose to replace model evaluation with a coin flip, solving this long standing problem, and saving a lot of ressources. We call it the ***Rademacher Coin Flipping*** metric.

### 4.1.2 Text

For text modeling we considered an autoregressive-bilstm-attention-cnn model. In the end we settled with the identity function due to the fact it is auto-invertible, which is nice.

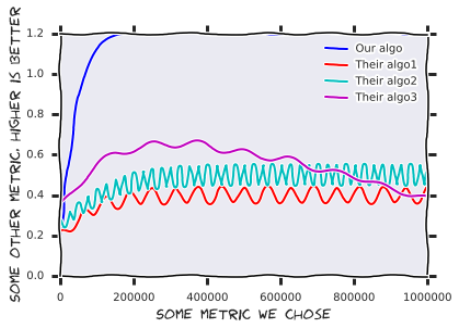

Figure 1: Generic results achieved by our algorithm. The advantage is that this figure can be reused for other papers.

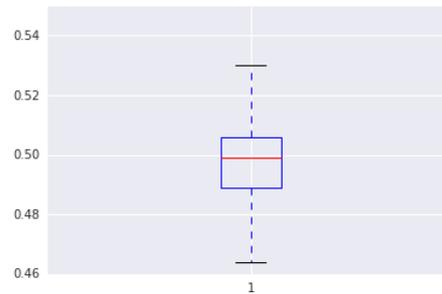

Figure 2: Human Evaluation. One sample from the test set and from the generative model are shown side-by-side to calibrated human raters.

Here is a representative sample of text generated by *MemGEN* trained on Wikipedia:

> April Fools' Day (sometimes called All Fools' Day) is an annual celebration in some European and Western countries commemorated on April 1 by playing practical jokes and spreading hoaxes. The jokes and their victims are called April fools. People playing April Fool jokes expose their prank by shouting "April fool". Some newspapers, magazines and other published media report fake stories, which are usually explained the next day or below the news section in small letters. Although popular since the 19th century, the day is not a public holiday in any country.

### 4.1.3 Generated Images

In Figure 3 we show some representative generated images of *MemGEN* after training on internet images. To the untrained eye it seems that the model *collapsed* and only generates cat pictures. Maybe the internet is full of cat images? A quick estimate of natural images found online seems to confirm this hypothesis, but further investigation is required.

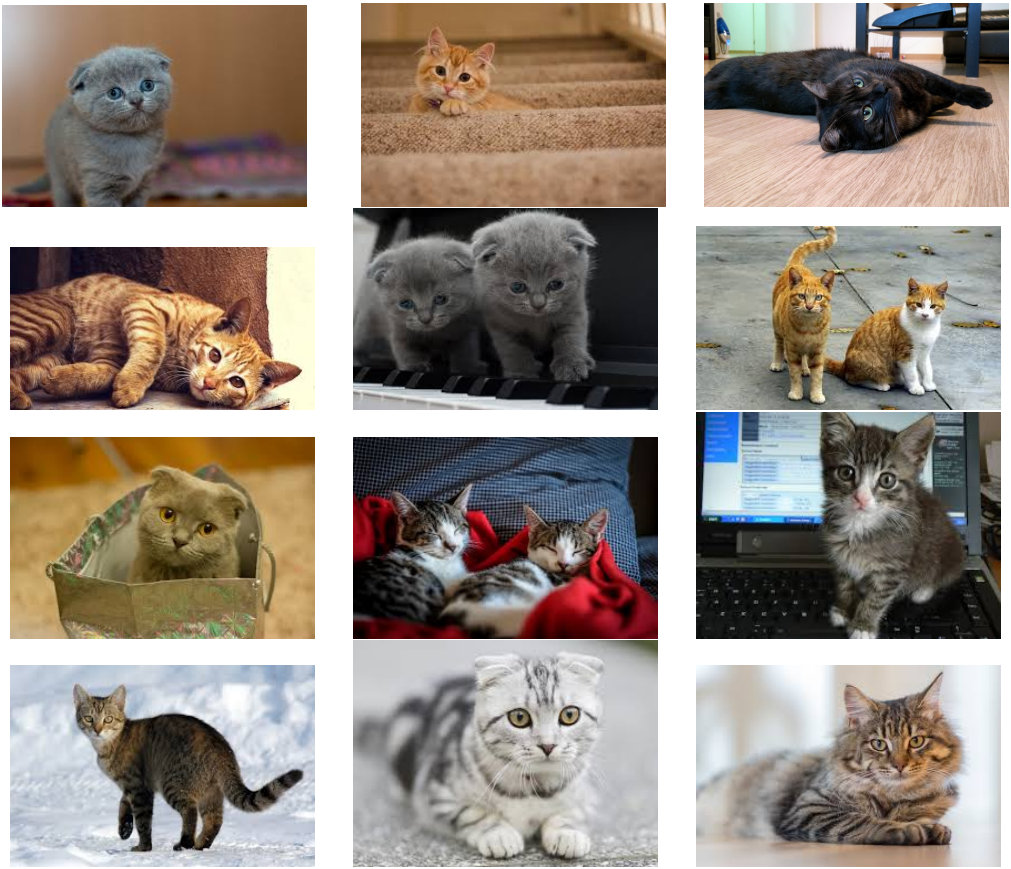

Figure 3: Random pictures generated by *MemGEN* after learning from random internet images. This is a proof that the generated data distribution is representative of the learning distribution.

### 4.1.4 PI digits

Here we show a random sample from a model trained on $\pi$, digits were produced as long as the memory permitted[5]. Please note that these are preliminary results, as we had a bug somewhere.

> 3.1415Segmentation fault

## 4.2 Computational Resources

We would like to carefully study the computational tradeoffs, and here we show that the results are outstanding, otherwise we would not have included them.

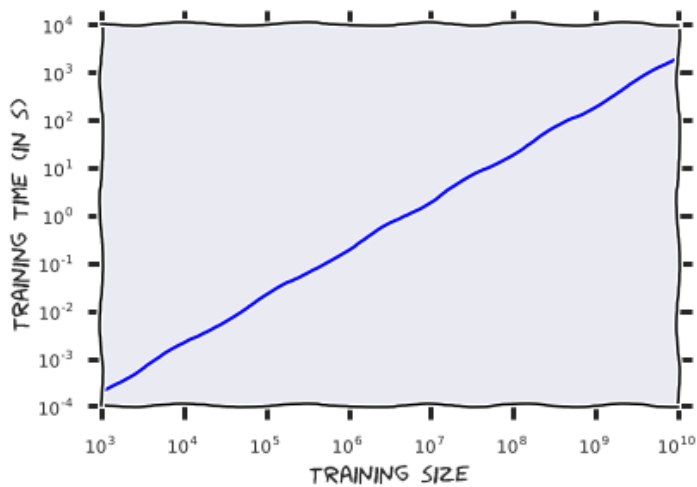

Figure 4: Training time of Algorithm 1 as a function of the dataset size on a single core of an old CPU.

Figure 4 shows that our algorithm significantly outperforms any previous reasonable generative model algorithm in terms of computational resources. We would also like to highlight that MemGEN can run on a simple CPU, hence reducing the training cost significantly. Folklore dictates that a custom FPGA implementation might provide a performance benefit.

However, GPUs have better calorific value than CPUs; due to the global GPU utilisation drop, Winter is coming.

---

[5]We assumed that 640 kB ought to be enough

# 5    Conclusions

We presented a novel generative modeling algorithm that has only advantages[6], has provably great properties and outstanding results on serious metrics. In contrast to [LaLoudouana and Tarare, 2003], we did not even have to select the dataset – we only selected the metrics. This paper is written following the best scientific principles, and any visible flaw is purely coincidental.

# Acknowledgements

Thanks to the Limbic System[7] team in Zurich and in other places with lower chocolate consumption, for insightful discussions leading to this breakthrough algorithm focusing on real problems.

---

[6] It actually has disadvantages, but we omit them for lack of space.
[7] Part of Brain